\pdfoutput=1

\documentclass[11pt]{article}

\usepackage[final]{acl}

\usepackage{times}
\usepackage{latexsym}
\usepackage{subcaption} 
\usepackage{xspace} 

\newcommand{\dtwolora}{D\textsuperscript{2}LoRA\xspace}

\usepackage{multirow}
\usepackage{booktabs,arydshln}
\usepackage{dcolumn}
\usepackage{array}
\usepackage{tabularx}
\usepackage{amsmath}
\usepackage{cleveref}
\usepackage{lipsum}
\usepackage{tcolorbox}
\crefname{subsection}{subsection}{subsections}
\usepackage{amssymb} 
\usepackage{enumitem} 
\usepackage[T1]{fontenc}

\usepackage[utf8]{inputenc}

\usepackage{microtype}

\usepackage{inconsolata}

\usepackage{graphicx}
\usepackage{todonotes}
\usepackage{booktabs}

\usepackage{cleveref}
\newcommand{\fon}[1]{\fontfamily{#1}\selectfont}

\title{\dtwolora: Data-Driven LoRA Initialization for Low Resource Tasks} 

\author{Author 1 \\ {\bf Author 2} \\ ... \\ {\bf Author n} \\
        Address line \\ ... \\ Address line}

\author{Javad Seraj \\ \textbf{Mohammad Mahdi Mohajeri} \\ \textbf{Mohammad Javad Dousti} \\
Department of Electrical and Computer Engineering \\ University of Tehran \\ 
\texttt{\{javad.seraj,mehdimohajeri,mjdousti\}@ut.ac.ir}}

\usepackage{tabularx}
\begin{document}
\maketitle

\begin{abstract}
Tuning large language models is essential for optimizing their performance across diverse applications, particularly in scenarios with limited data availability.Tuning large language models in scarce data scenarios is crucial, particularly given that the convergence speed of the LoRA method is lower than that of full fine-tuning. In this paper, we present an analysis of post-training methods—including Supervised Fine-Tuning (SFT), Direct Preference Optimization (DPO), and Odds Ratio Preference Optimization (ORPO)—within the context of task-specific learning using the LoRA method. Next we introduce \dtwolora, a data-driven approach for initializing LoRA metrics that enhances training efficiency, especially in limited-data settings. Our experiments compare \dtwolora with vanilla LoRA in terms of performance and catastrophic forgetting under extremely data-constrained conditions. The results demonstrate that \dtwolora achieves a 1\% improvement GSM8K benchmark and a 2-point improvement in ROUGE score in title generation tasks. \dtwolora facilitates the adaptation of LLMs to multiple tasks even when task-specific data is scarce, thereby reducing training expenses and offering data cost.
\end{abstract}

\section{Introduction}
Large Language Models (LLMs) have demonstrated superior performance in various tasks.
%
For general improvement or downstream tasks, they are fine-tuned on domain-specific data or aligned with preference data to human preferences~\cite{ouyang2022instructGPT, rafailov2024DPO}.
Increasing the number of parameters in LLMs or low availability of training data causes full fine-tuning methods to become challenging~\cite{kaddour2023challenges, raiaan2024review}.
A popular solution to the large number of training parameters is the use of parameter-efficient fine-tuning (PEFT) methods.  
PEFT methods reduce fine-tuning parameters and memory usage.
Low-rank adaptors (LoRA) is one of the most popular PEFT methods~\cite{hu2021lora}.

The LoRA method trains two low-rank matrices, namely $A$ and $B$, for each layer during fine-tuning while the model's pre-trained weights ($W$) are frozen~\cite{hu2021lora}.

Using the two low-rank matrices, another matrix called $\Delta W$ is approximated as $\Delta W=A \times B$.
The final weight matrix is then obtained by adding the low-rank approximation to the original pre-trained weights ($W+\Delta W$).

The initialization of matrices in the LoRA method impacts its outcome quality~\cite{hayou2024impact, mao2025survey}.
The original LoRA method initializes one of the low-rank matrices using a Gaussian distribution, while the other matrix is set to zero.~\cite{hu2021lora}.
MULTILoRA~\cite{wang2023multilora} uses Kaiming-uniform~\cite{he2015kaiming} to initialize one of matrices.
%
%
MiLoRA~\cite{wang2024milora} initializes low-rank matrices in a subspace orthogonal to the principal matrix (i.e., $W$), which helps ensure the preservation of pre-trained knowledge.
PiSSA~\cite{meng2024pissa} follows the LoRA architecture but initializes $A$ and $B$ with the principal components of $W$.
The remaining components are stored in a frozen residual matrix $W^{res}$ with the same size of $\Delta W$.
This approach updates only the principal components, enabling faster convergence and better performance.

OLoRA~\cite{buyukakyuz2024olora} enhances the LoRA method by employing orthonormal matrix initialization via QR decomposition.
This approach accelerates LLM training convergence while maintaining LoRA's efficiency in terms of trainable parameters and GPU memory usage.
CorDA~\cite{yangcorda_corda} leverages singular value decomposition of pre-trained weights, guided by a covariance matrix representing task-specific context.
By aggregating contextual information into principal components, CorDA supports two adaptation modes:
knowledge-preserved adaptation, which mitigates world knowledge forgetting, and instruction-previewed adaptation, which enhances fine-tuning performance.
%

In this paper, we introduce a \textit{Data-Driven approach for LoRA initialization}. We call our approach \dtwolora.
Our approach uses general and high-quality data for initializing LoRA matrices.
In our approach, training is conducted in two phases.
In the first phase, called \textit{warm-up}, LoRA matrices are trained using a general high-quality dataset.
In the second phase, \textit{task adaptation}, the initialized matrices are used in a standard manner for fine-tuning or preference optimizing on downstream tasks. 
Although our approach differs from prior works that initialize LoRA matrices based on the pre-trained model's weights, it remains compatible with previous initialization methods. 
%
%
Our approach significantly reduces the need for in-domain and task-specific samples to perform fine-tuning.


%
The rest of this paper is organized as follows.
In \Cref{sec:methodology}, we explain our methodology and formally formulate our method. 
In \Cref{sec:setup}, we present the dataset used and our experimental setup.
Finally, we discuss the results and provide an analysis.  

\section{Data-Driven LoRA Initialization}
\label{sec:methodology}
Fine-tuning of LLMs in downstream tasks requires a large number of data to achieve the desired performance.
In limited-data scenarios, model fine-tuning faces significant challenges.  
Due to the lower number of optimization steps in such cases, LLMs may struggle to learn tasks effectively.  
Moreover, increasing the number of fine-tuning steps on limited data can lead to overfitting.  
The challenge of fine-tuning in tasks with limited data becomes particularly significant when employing the LoRA method.
As mentioned in PiSSA~\cite{meng2024pissa}, LoRA exhibits slower convergence compared to full fine-tuning.

We introduce a data-driven initialization method for LoRA, named \dtwolora, which leverages high-quality general data to enhance performance in limited-data tasks.  

\dtwolora \space consists of two phases. In the first phase, called the warm-up phase, general high-quality data is used to initialize LoRA weights ($\Delta W$), facilitating faster convergence.  
The second phase, called the task adaptation phase, follows the standard LoRA training procedure.  
In this phase, task-specific data is used to adapt the initialized LoRA matrices to the target task.  

\subsection{Mathematical Formulation}

\dtwolora consists of two distinct phases: the warm-up phase and the task adaptation phase.  
During the warm-up phase, we use $m$ high-quality general data samples to train low-rank matrices.  
In the subsequent task adaptation phase, we fine-tune the initialized matrices using $n$ in-domain data samples.  
We define the model's final weights as $W_{\dtwolora(m, n)}$, representing the weights obtained after training with $m$ general data samples in the warm-up phase and $n$ task-specific samples in the adaptation phase.  
Both training phases follow a methodology similar to the standard LoRA~\cite{hu2021lora}, such that:  
$$W_{D^2LoRA(m, n)} = W + \Delta W_{m,n}$$  
Under this formulation, the weights of vanilla LoRA, denoted as $W_{LoRA}$, can be expressed as $W_{D^2LoRA(0, n)}$,  
since vanilla LoRA is only fine-tuned on $n$ in-domain data samples.  
This indicates that \dtwolora extends vanilla LoRA by incorporating an additional training phase with $m$ general data samples.  

To compare vanilla LoRA and \dtwolora, we evaluate their performance on specific tasks.  
We define the performance of a model with weights $W_{model}$ on a given task $t$ as $Perf(W_{model}, t)$.  
When the number of in-domain data samples $n$ is small, we expect that \dtwolora will outperform vanilla LoRA:  
%
$$Perf(W_{\dtwolora(m, n)}, t) > Perf(W_{LoRA}, t)$$  
However, as $n$ increases, we anticipate that the performance of both methods will be the same:  
%
$$Perf(W_{\dtwolora(m, n)}, t) \approx Perf(W_{LoRA}, t)$$  
This suggests that the primary advantage of $D^2LoRA$ emerges in scenarios where the available in-domain data is limited.  

\section{Experimental Setup}
\label{sec:setup}
We choose Llama3.1-8B-Instruct~\cite{dubey2024llama3} as the base model. For training, we employ PEFT techniques, specifically LoRA modules.
%

Our research is based on two main experiments. In the first experiment, we analyze the impact of LoRA matrices across different fine-tuning methods, including Supervised Fine-Tuning (SFT), Direct Preference Optimization (DPO)~\cite{rafailov2024DPO}, and Odds Ratio Preference Optimization (ORPO)~\cite{hong-etal-2024-orpo}. We progressively increase the training data and examine the learning curve of each method. Details and results are provided in ~\Cref{subsec:exp1}. By understanding LoRA's behavior across different post-training methods, this experiment establishes a foundation for comparing \dtwolora with LoRA in subsequent experiments.

In the second experiment, we explore our data-driven approach to initializing LoRA matrices and assess its impact on the training process, particularly in data-limited scenarios. The analysis and results are presented in ~\Cref{subsec:exp2}. 

Additionally, we conduct two further experiments to deepen our analysis of \dtwolora{}’s effectiveness. The first investigates catastrophic forgetting~\cite{luo2023cf} in LoRA compared to \dtwolora, and the second extends our method to extremely limited-data scenarios. Details of these experiments are provided in ~\Cref{subsec:cf,subsec:less1000}.

%

\subsection{Datasets}
\label{sec:dataset}
%

Our experiments were conducted on two tasks, namely mathematical reasoning and title generation. For mathematical reasoning, we used the \textit{Math-Step-DPO-10K}~\cite{lai2024step} dataset (described in ~\Cref{appendix:math_dataset}), which contains 10,000 step-by-step mathematical reasoning preference samples. For title generation, we employed the \textit{Economics Research Paper}~\cite{econ} dataset (detailed in ~\Cref{appendix:title_gen}), comprising 7,070 instruction samples, each with an abstract and corresponding title. For the warm-up step, we used instructions and chosen answers from the \textit{40K-Mix}\footnote{Available at \url{https://huggingface.co/datasets/mlabonne/orpo-dpo-mix-40k}.} dataset, a combination of several high-quality preference optimization datasets. Details are provided in ~\Cref{appendix:general_dataset}.

\subsection{Evaluation Setup}
\label{sec:eval_setup}
For mathematical reasoning, we evaluate \dtwolora on the GSM8K~\cite{cobbe2021gsm8k} benchmark in a zero-shot setting, measuring performance based on the accuracy of the final answer.  
For title generation, we evaluate using 100 abstract-title pairs, with the ROUGE metrics between the generated title and the ground truth serving as the performance measure.

\subsection{Experiment 1: Impact of Data Scaling on LoRA’s Performance}
\label{subsec:exp1}

In this experiment, we evaluated the data efficiency of three post-training methods—ORPO, DPO, and SFT—when applying LoRA modules. To achieve this, the models were trained with varying dataset sizes, starting from 1,000 instances and gradually increasing to 2,000, 5,000, and finally 10,000 instances in mathematical reasoning.  

This incremental increase in dataset size allowed us to analyze the effect of the number of instances on model performance convergence, providing insights into the sample efficiency of each post-training method. Additionally, this experiment identified the convergence point for each post-training method, where further increasing the dataset size results in a stable performance without significant improvements. 

\begin{figure}[ht]
    \centering
    \includegraphics[width=0.95\linewidth]{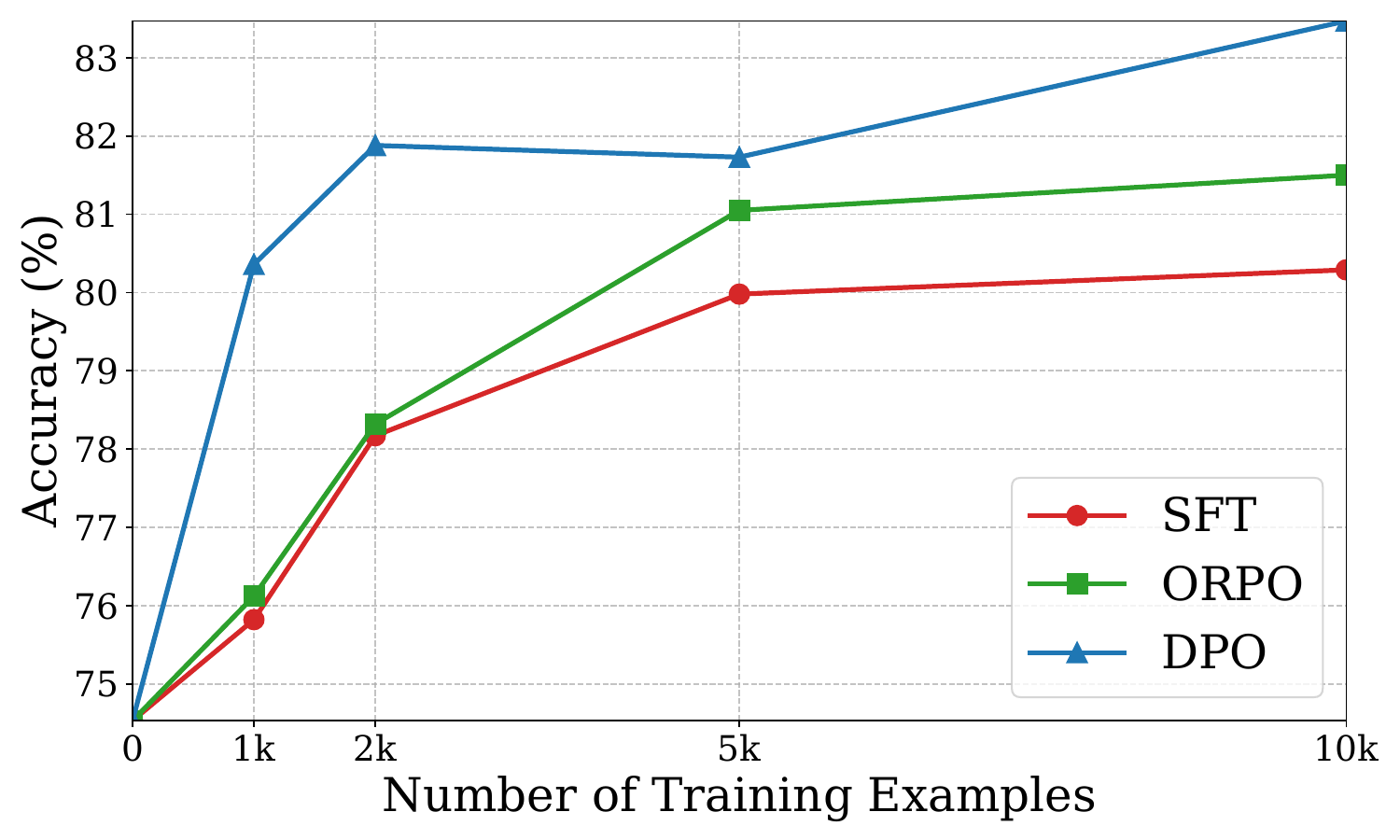}
    \caption{Performance comparison of three training methods (SFT, DPO, ORPO) on the GSM8K benchmark}
    \label{fig:gsm_methods_compare}
\end{figure}

As shown in ~\Cref{fig:gsm_methods_compare}, ORPO and SFT exhibit similar performance in the limited-data scenarios. However, as the number of instances increases, ORPO surpasses SFT, showing more significant performance gains. Notably, DPO significantly outperforms both SFT and ORPO in limited-data scenarios.

\subsection{Experiment 2: \dtwolora\ Effectiveness}
\label{subsec:exp2}

\begin{figure}[ht]
    \centering
    \includegraphics[width=0.95\linewidth]{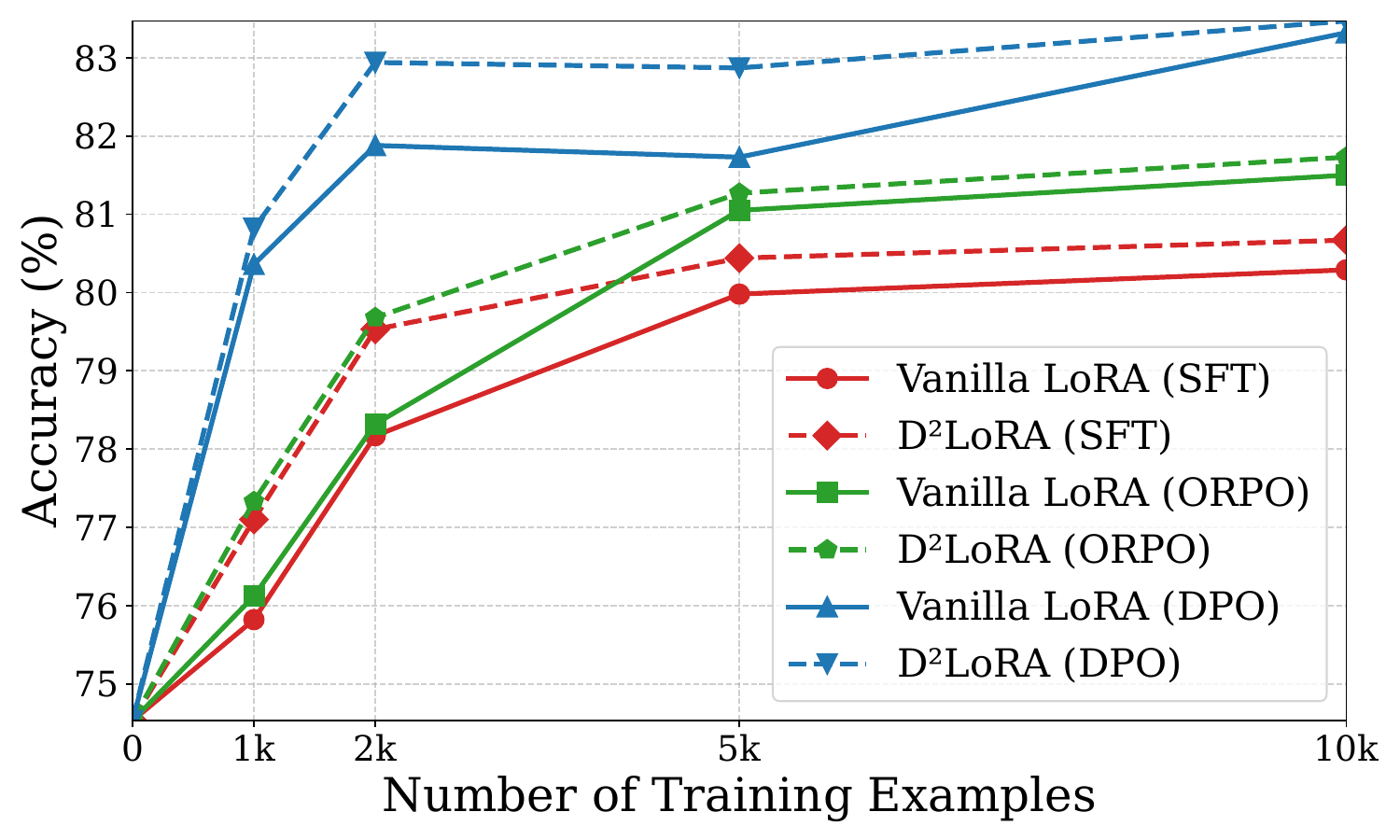}
    \caption{Comparison of \dtwolora and vanilla LoRA across different training methods (SFT, DPO, ORPO) on the GSM8K benchmark.}
    \label{fig:effectiveness}
\end{figure}

In the second experiment, we examined the impact of initializing LoRA matrices using a preparatory warm-up step.
This step involved training the LoRA matrices on a high-quality general instruction-tuning dataset. Specifically, we utilized the instructions and chosen responses from the 40K-Mix dataset which detailed in ~\Cref{appendix:dataset_info}. 

\begin{table}[ht]
    \centering
    \small 
    \resizebox{\columnwidth}{!}{
    \begin{tabular}{lccccc}
        \toprule
        \multirow{2}{*}{\textbf{Training Method}} & \multicolumn{5}{c}{\textbf{Number of Training Instances}} \\ 
        \cmidrule(lr){2-6}
        & \textbf{0} & \textbf{1k} & \textbf{2k} & \textbf{5k} & \textbf{7k} \\
        \midrule
        SFT                  & 21.62 & 24.81 & 27.23 & 29.80 & 29.97 \\
        SFT (\dtwolora{}(40k,n)) & 22.19 & 26.90 & 28.10 & 30.03 & 31.66 \\
        \bottomrule
    \end{tabular}
    }
    \caption{Comparing ROUGE Score of LoRA and \dtwolora in the Title Generation Task.}
    \label{tab:orpo_accuracytitle_generation_performance}
\end{table}

After initializing the LoRA matrices in the warm-up step, we proceed with experiments on SFT, DPO, and ORPO using these initialized matrices to assess the overall effectiveness of \dtwolora. The results, presented in ~\Cref{fig:effectiveness} and ~\Cref{tab:d2lora_math_results}, highlight the improvements in task performance achieved by incorporating \dtwolora into mathematical reasoning. 

This improvement is evident across all three post-training methods. As shown in ~\Cref{fig:effectiveness}, the effect is particularly pronounced in the limited-data regime. However, as the number of training instances increases, the performance gap between vanilla LoRA and \dtwolora narrows, as observed in~\Cref{fig:effectiveness}.  Furthermore, ~\Cref{tab:orpo_accuracytitle_generation_performance} presents the improvements in ROUGE scores for the title generation task, demonstrating the effectiveness of the \dtwolora against LoRA on title generation tasks.

\subsection{Experiment 3: Impact of Task-Specific Training on General Kwonldge of LLMs}
\label{subsec:cf}

\begin{figure}[ht]
    \centering
    \includegraphics[width=0.95\linewidth]{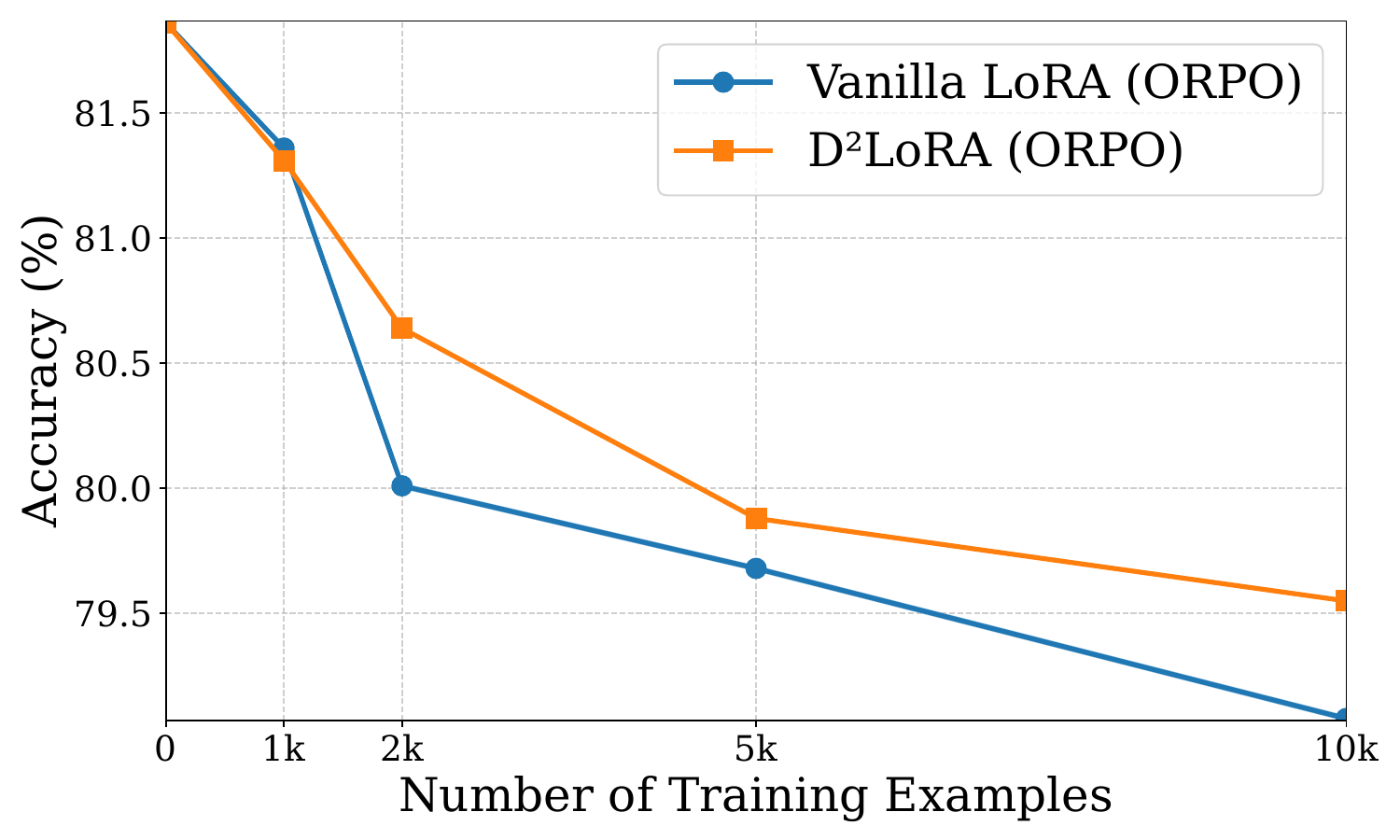}
    \caption{Impact of task-specific training on general knowledge and reasoning accuracy on the ARC Benchmark. As task-specific training data increases, the original model's general reasoning ability declines.}
    \label{fig:forgetting}
\end{figure}

We compare the \dtwolora method against vanilla LoRA in addressing the challenges of catastrophic forgetting by increasing the number of training instances. For this experiment, we evaluated on the ARC benchmark~\cite{arc2018}, which consists of 2,378 exam questions detailed in~\Cref{sec:arc_easy}. The results, presented in ~\Cref{fig:forgetting}, show that as the number of training instance for the target task increases, performance on the ARC benchmark declines. However, we observe no significant difference in the accuracy drop between vanilla LoRA and \dtwolora, suggesting that \dtwolora does not exacerbate catastrophic forgetting.

%
%
%


\subsection{Experiment 4: Analyzing \dtwolora in Extremely Limited Data Scenarios}
\label{subsec:less1000}

In this experiment, we examine the effectiveness of the \dtwolora method in scenarios with extremely scarce data. Specifically, we focus on cases with fewer than 1,000 instances.\Cref{fig:less_1000} presents the results of training ORPO with both \dtwolora and vanilla LoRA. The results show that \dtwolora outperforms vanilla LoRA in extremely scarce data scenarios.

\section{Conclusion}
In this work, we introduce \dtwolora, a data-driven approach for initializing LoRA. We analyze post-training methods (SFT, DPO, ORPO) in the context of learning a specific task using LoRA modules, examining the behavior of each method as the number of training instances increases. This experiment provides deeper insights into \dtwolora{}'s behavior in limited-data scenarios, enhancing training efficiency in such settings. We further compare \dtwolora with vanilla LoRA in terms of catastrophic forgetting and performance in extremely data-limited conditions. The \dtwolora approach achieves a 1\% improvement in mathematical reasoning and a 2-point increase in ROUGE score for title generation. Additionally, \dtwolora not only enhances model performance in data-scarce scenarios but also improves cost efficiency by reducing training expenses. By utilizing a single initialized LoRA module, \dtwolora enables the adaptation of LLMs to multiple tasks.

\section{Limitations} 
Our evaluation of \dtwolora's impact is limited to Llama3.1-8B. Further evaluations across diverse model families and sizes are necessary to comprehensively assess its effectiveness. \dtwolora is a compatible method with parameter-efficient adapter techniques and proves effective when task-specific data is scarce. However, as the amount of task-specific training data increases, the effectiveness of \dtwolora diminishes. We utilized a specific dataset with a fixed number of data points during the warm-up phase. To gain a more comprehensive understanding of this phase, further evaluations using diverse datasets and varying data quantities are necessary.

\bibliography{acl_latex}

\begin{thebibliography}{20}
\providecommand{\natexlab}[1]{#1}

\bibitem[{B{\"u}y{\"u}kaky{\"u}z(2024)}]{buyukakyuz2024olora}
Kerim B{\"u}y{\"u}kaky{\"u}z. 2024.
\newblock Olora: Orthonormal low-rank adaptation of large language models.
\newblock \emph{arXiv preprint arXiv:2406.01775}.

\bibitem[{Clark et~al.(2018)Clark, Cowhey, Etzioni, Khot, Sabharwal, Schoenick, and Tafjord}]{arc2018}
Peter Clark, Isaac Cowhey, Oren Etzioni, Tushar Khot, Ashish Sabharwal, Carissa Schoenick, and Oyvind Tafjord. 2018.
\newblock Think you have solved question answering? try arc, the ai2 reasoning challenge.
\newblock \emph{ArXiv}, abs/1803.05457.

\bibitem[{Cobbe et~al.(2021)Cobbe, Kosaraju, Bavarian, Chen, Jun, Kaiser, Plappert, Tworek, Hilton, Nakano, Hesse, and Schulman}]{cobbe2021gsm8k}
Karl Cobbe, Vineet Kosaraju, Mohammad Bavarian, Mark Chen, Heewoo Jun, Lukasz Kaiser, Matthias Plappert, Jerry Tworek, Jacob Hilton, Reiichiro Nakano, Christopher Hesse, and John Schulman. 2021.
\newblock Training verifiers to solve math word problems.
\newblock \emph{arXiv preprint arXiv:2110.14168}.

\bibitem[{Dubey et~al.(2024)Dubey, Jauhri, Pandey, Kadian, Al-Dahle, Letman, Mathur, Schelten, Yang, Fan et~al.}]{dubey2024llama3}
Abhimanyu Dubey, Abhinav Jauhri, Abhinav Pandey, Abhishek Kadian, Ahmad Al-Dahle, Aiesha Letman, Akhil Mathur, Alan Schelten, Amy Yang, Angela Fan, et~al. 2024.
\newblock The llama 3 herd of models.
\newblock \emph{arXiv preprint arXiv:2407.21783}.

\bibitem[{Hayou et~al.(2024)Hayou, Ghosh, and Yu}]{hayou2024impact}
Soufiane Hayou, Nikhil Ghosh, and Bin Yu. 2024.
\newblock The impact of initialization on lora finetuning dynamics.
\newblock \emph{arXiv preprint arXiv:2406.08447}.

\bibitem[{He et~al.(2015)He, Zhang, Ren, and Sun}]{he2015kaiming}
Kaiming He, Xiangyu Zhang, Shaoqing Ren, and Jian Sun. 2015.
\newblock Delving deep into rectifiers: Surpassing human-level performance on imagenet classification.
\newblock In \emph{Proceedings of the IEEE International Conference on Computer Vision (ICCV)}, pages 1026--1034.

\bibitem[{Hong et~al.(2024)Hong, Lee, and Thorne}]{hong-etal-2024-orpo}
Jiwoo Hong, Noah Lee, and James Thorne. 2024.
\newblock \href {https://doi.org/10.18653/v1/2024.emnlp-main.626} {{ORPO}: Monolithic preference optimization without reference model}.
\newblock In \emph{Proceedings of the 2024 Conference on Empirical Methods in Natural Language Processing}, pages 11170--11189, Miami, Florida, USA. Association for Computational Linguistics.

\bibitem[{Hu et~al.(2021)Hu, Shen, Wallis, Allen-Zhu, Li, Wang, Wang, and Chen}]{hu2021lora}
Edward~J Hu, Yelong Shen, Phillip Wallis, Zeyuan Allen-Zhu, Yuanzhi Li, Shean Wang, Lu~Wang, and Weizhu Chen. 2021.
\newblock Lora: Low-rank adaptation of large language models.
\newblock \emph{arXiv preprint arXiv:2106.09685}.

\bibitem[{Kaddour et~al.(2023)Kaddour, Harris, Mozes, Bradley, Raileanu, and McHardy}]{kaddour2023challenges}
Jean Kaddour, Joshua Harris, Maximilian Mozes, Herbie Bradley, Roberta Raileanu, and Robert McHardy. 2023.
\newblock Challenges and applications of large language models.
\newblock \emph{arXiv preprint arXiv:2307.10169}.

\bibitem[{Keles and Bayrakl{\i}(2024)}]{econ}
Onur Keles and Omer~Turan Bayrakl{\i}. 2024.
\newblock \href {https://aclanthology.org/2024.finnlp-1.21/} {{LL}a{MA}-2-econ: Enhancing title generation, abstract classification, and academic {Q}{\&}{A} in economic research}.
\newblock In \emph{Proceedings of the Joint Workshop of the 7th Financial Technology and Natural Language Processing, the 5th Knowledge Discovery from Unstructured Data in Financial Services, and the 4th Workshop on Economics and Natural Language Processing}, pages 212--218, Torino, Italia. Association for Computational Linguistics.

\bibitem[{Lai et~al.(2024)Lai, Tian, Chen, Yang, Peng, and Jia}]{lai2024step}
Xin Lai, Zhuotao Tian, Yukang Chen, Senqiao Yang, Xiangru Peng, and Jiaya Jia. 2024.
\newblock Step-dpo: Step-wise preference optimization for long-chain reasoning of llms.
\newblock \emph{arXiv preprint arXiv:2406.18629}.

\bibitem[{Luo et~al.(2023)Luo, Yang, Meng, Li, Zhou, and Zhang}]{luo2023cf}
Yun Luo, Zhen Yang, Fandong Meng, Yafu Li, Jie Zhou, and Yue Zhang. 2023.
\newblock An empirical study of catastrophic forgetting in large language models during continual fine-tuning, 2023.
\newblock \emph{URL https://arxiv. org/abs/2308.08747}.

\bibitem[{Mao et~al.(2025)Mao, Ge, Fan, Xu, Mi, Hu, and Gao}]{mao2025survey}
Yuren Mao, Yuhang Ge, Yijiang Fan, Wenyi Xu, Yu~Mi, Zhonghao Hu, and Yunjun Gao. 2025.
\newblock A survey on lora of large language models.
\newblock \emph{Frontiers of Computer Science}, 19(7):197605.

\bibitem[{Meng et~al.(2024)Meng, Wang, and Zhang}]{meng2024pissa}
Fanxu Meng, Zhaohui Wang, and Muhan Zhang. 2024.
\newblock Pissa: Principal singular values and singular vectors adaptation of large language models.
\newblock \emph{arXiv preprint arXiv:2404.02948}.

\bibitem[{Ouyang et~al.(2022)Ouyang, Wu, Jiang, Almeida, Wainwright, Mishkin, Zhang, Agarwal, Slama, Ray et~al.}]{ouyang2022instructGPT}
Long Ouyang, Jeffrey Wu, Xu~Jiang, Diogo Almeida, Carroll Wainwright, Pamela Mishkin, Chong Zhang, Sandhini Agarwal, Katarina Slama, Alex Ray, et~al. 2022.
\newblock Training language models to follow instructions with human feedback.
\newblock \emph{Advances in neural information processing systems}, 35:27730--27744.

\bibitem[{Rafailov et~al.(2024)Rafailov, Sharma, Mitchell, Manning, Ermon, and Finn}]{rafailov2024DPO}
Rafael Rafailov, Archit Sharma, Eric Mitchell, Christopher~D Manning, Stefano Ermon, and Chelsea Finn. 2024.
\newblock Direct preference optimization: Your language model is secretly a reward model.
\newblock \emph{Advances in Neural Information Processing Systems}, 36.

\bibitem[{Raiaan et~al.(2024)Raiaan, Mukta, Fatema, Fahad, Sakib, Mim, Ahmad, Ali, and Azam}]{raiaan2024review}
Mohaimenul Azam~Khan Raiaan, Md~Saddam~Hossain Mukta, Kaniz Fatema, Nur~Mohammad Fahad, Sadman Sakib, Most Marufatul~Jannat Mim, Jubaer Ahmad, Mohammed~Eunus Ali, and Sami Azam. 2024.
\newblock A review on large language models: Architectures, applications, taxonomies, open issues and challenges.
\newblock \emph{IEEE Access}.

\bibitem[{Wang et~al.(2024)Wang, Li, Wang, Chen, and Chen}]{wang2024milora}
Hanqing Wang, Yixia Li, Shuo Wang, Guanhua Chen, and Yun Chen. 2024.
\newblock Milora: Harnessing minor singular components for parameter-efficient llm finetuning.
\newblock \emph{arXiv preprint arXiv:2406.09044}.

\bibitem[{Wang et~al.(2023)Wang, Lin, Zeng, and Zhang}]{wang2023multilora}
Yiming Wang, Yu~Lin, Xiaodong Zeng, and Guannan Zhang. 2023.
\newblock Multilora: Democratizing lora for better multi-task learning.
\newblock \emph{arXiv preprint arXiv:2311.11501}.

\bibitem[{Yang et~al.(2024)Yang, Li, Zhou, Song, Wu, Nie, and Ghanem}]{yangcorda_corda}
Yibo Yang, Xiaojie Li, Zhongzhu Zhou, Shuaiwen~Leon Song, Jianlong Wu, Liqiang Nie, and Bernard Ghanem. 2024.
\newblock Corda: Context-oriented decomposition adaptation of large language models for task-aware parameter-efficient fine-tuning.
\newblock In \emph{The Thirty-eighth Annual Conference on Neural Information Processing Systems}.

\end{thebibliography}
\clearpage

\appendix

\section{Prompts and Dataset Examples}
In this section, we present prompts and examples from the training datasets.
we present the prompts used for inference with the model on the GSM8K benchmark, chain of thought in the Math-Step-DPO dataset, and instance from the Economic Paper Title Generation dataset.

\begin{tcolorbox}[
fonttitle=\small\fon{pbk}\bfseries,
    left=2pt,
    right=2pt,
    top=2pt,
    bottom=2pt,
    title=GSM8K Inference Prompt,
]
\textbf{User:} \{instruction\}

Please reason step by step, and put your final answer within boxed\{\{\}\}.
\\ \\
\textbf{Assistant:}
\end{tcolorbox}

\begin{tcolorbox}[
fonttitle=\small\fon{pbk}\bfseries,
    left=2pt,
    right=2pt,
    top=2pt,
    bottom=2pt,
    title=Example of the Initial Chain of Thought in the Math-Step-DPO Dataset,
]
\textbf{prompt:}\\
If Grace charges 300 dollars per week and her client pays her every 2 weeks, how many weeks will it take for Grace to earn a total of 1800 dollars?

\textbf{initial\_reason\_steps:}\\
Let's think step by step. Step 1: First, we need to figure out how much Grace earns in 2 weeks. Since she charges 300 dollars per week, in 2 weeks she earns $300 \times 2 = 600$ dollars. Step 2: Now, we want to find out how many 2-week periods it takes for Grace to earn 1800 dollars. To do this, we divide the total amount she wants to earn by the amount she earns in each 2-week period. So, we divide 1800 by 600. Step 3:

\end{tcolorbox}

\begin{tcolorbox}[
fonttitle=\small\fon{pbk}\bfseries,
    left=2pt,
    right=2pt,
    top=2pt,
    bottom=2pt,
    title=Example of an Instance from the Economic Paper Title Generation Dataset,
]
\textbf{text:}\\
In this paper, we propose a model which simulates odds distributions of pari-mutuel betting system under two hypotheses on the behavior of bettors: 1. The amount of bets increases very rapidly as the deadline for betting comes near. 2. Each bettor bets on a horse which gives the largest expectation value of the benefit. The results can be interpreted as such efficient behaviors do not serve to extinguish the FL bias but even produce stronger FL bias.
\\ \\
\textbf{title:}\\
Efficiency in Micro-Behaviors and FL Bias

\end{tcolorbox}

\section{Training Details}
\label{appendix:training_details}
In ~\Cref{tab:training_hp}, the training parameters for each stage are presented for mathematical reasoning, title generation, and the initialization step. 
\begin{table*}[ht]
    \centering
    \small 
    \resizebox{\textwidth}{!}{
    \begin{tabular}{lcccccccccccccc}
        \toprule
        \multirow{3}{*}{\textbf{Task}} & \multirow{3}{*}{\textbf{Phase}} & \multirow{3}{*}{\textbf{Dataset}} & 
        \multirow{3}{*}{\textbf{Lora Rank}} & \multirow{3}{*}{\textbf{Lora Alpha}} & 
        \multicolumn{2}{c}{\textbf{Warmup (40K Mix)}} & 
        \multicolumn{8}{c}{\textbf{Task-Specific Training}} \\
        \cmidrule(lr){6-7} \cmidrule(lr){8-15}
        & & & & & \textbf{lr} & \textbf{epochs} & 
        \multicolumn{2}{c}{\textbf{SFT}} & \multicolumn{3}{c}{\textbf{DPO}} & \multicolumn{3}{c}{\textbf{ORPO}} \\
        \cmidrule(lr){8-9} \cmidrule(lr){10-12} \cmidrule(lr){13-15}
        & & & & & & & \textbf{lr} & \textbf{epochs} & \textbf{lr} & \textbf{epochs} & \textbf{beta} & \textbf{lr} & \textbf{epochs} & \textbf{beta} \\
        \midrule
        \multirow{2}{*}{Mathematical Reasoning} & Warmup & 40K Mix & 16 & 16 & 1e-7 & 1 & - & - & - & - & - & - & - & - \\
        & Task-Specific & Math-Step-DPO-10K & 16 & 16 & - & - & 1e-6 & 3 & 1e-6 & 4 & 0.1 & 1e-6 & 4 & 0.1 \\
        \midrule
        \multirow{2}{*}{Title Generation} & Warmup & 40K Mix & 16 & 16 & 1e-7 & 1 & - & - & - & - & - & - & - & - \\
        & Task-Specific & Econ-Research-7K & 16 & 16 & - & - & 1e-6 & 3 & N/A & N/A & N/A & N/A & N/A & N/A \\
        \bottomrule
    \end{tabular}
    }
    \caption{Training hyperparameters for mathematical reasoning and title generation tasks. Phase 1 is a warmup with the 40K Mix dataset, and Phase 2 is task-specific training with SFT, DPO, and ORPO methods. All training phases use a cosine learning scheduler. The beta hyperparameter for DPO and ORPO is set to 0.1. "N/A" indicates that there are no rejected answers for preference optimization in the title generation task.}
    \label{tab:training_hp}
\end{table*}

\section{Training Dataset and Benchmarks Information}
\label{appendix:dataset_info}

\subsection{Mathematical Reasoning Dataset}
\label{appendix:math_dataset}
\textbf{Mathematical Reasoning:} The Math-Step-DPO-10K~\cite{lai2024step} dataset comprises 10,795 high-quality, stepwise preference pairs for mathematical reasoning. This dataset was constructed to enhance the robustness and factuality of Large Language Models (LLMs) in long-chain reasoning tasks.

\subsection{Title Generation Dataset}
\label{appendix:title_gen}
\textbf{Title Generation:} The Economics Research Paper~\cite{econ} Dataset is designed for title generation and related NLP tasks. It comprises 6,362 economics research paper abstracts and titles, collected via the arXiv API from specified economics categories.

\subsection{General Initialization Dataset}
\label{appendix:general_dataset}
\textbf{Warmup Phase:} The DPO-ORPP-Mix-40k dataset is a curated combination of several high-quality Direct Preference Optimization (DPO) datasets, totaling 40,218 preference pairs. The dataset includes 7,424 samples from the Capybara Preferences dataset, 2,299 samples from the Distilabel Intel Orca DPO Pairs, 22,799 samples from the Ultrafeedback Binarized Preferences, 2,181 samples from the Distilabel Math Preference DPO, 541 samples from the Toxic DPO, 7,958 samples from the PRM DPO Pairs Cleaned, and 1,016 samples from the Truthy DPO. This dataset is designed to enhance alignment using high-quality preference data across diverse domains.

\subsection{ARC Easy Benchmark}
\label{sec:arc_easy}

The ARC Easy benchmark~\cite{arc2018} comprises 2,378 exam questions spanning several grade levels. Each question is structured as a multiple-choice item, typically offering four answer options.

\section{Additional Experiment Results}
\label{appendix:experiment_results}

In this section, we provide additional results from our experiments that demonstrate the effectiveness of the \dtwolora method.

\begin{figure}[ht]
    \centering
    \includegraphics[width=0.95\linewidth]{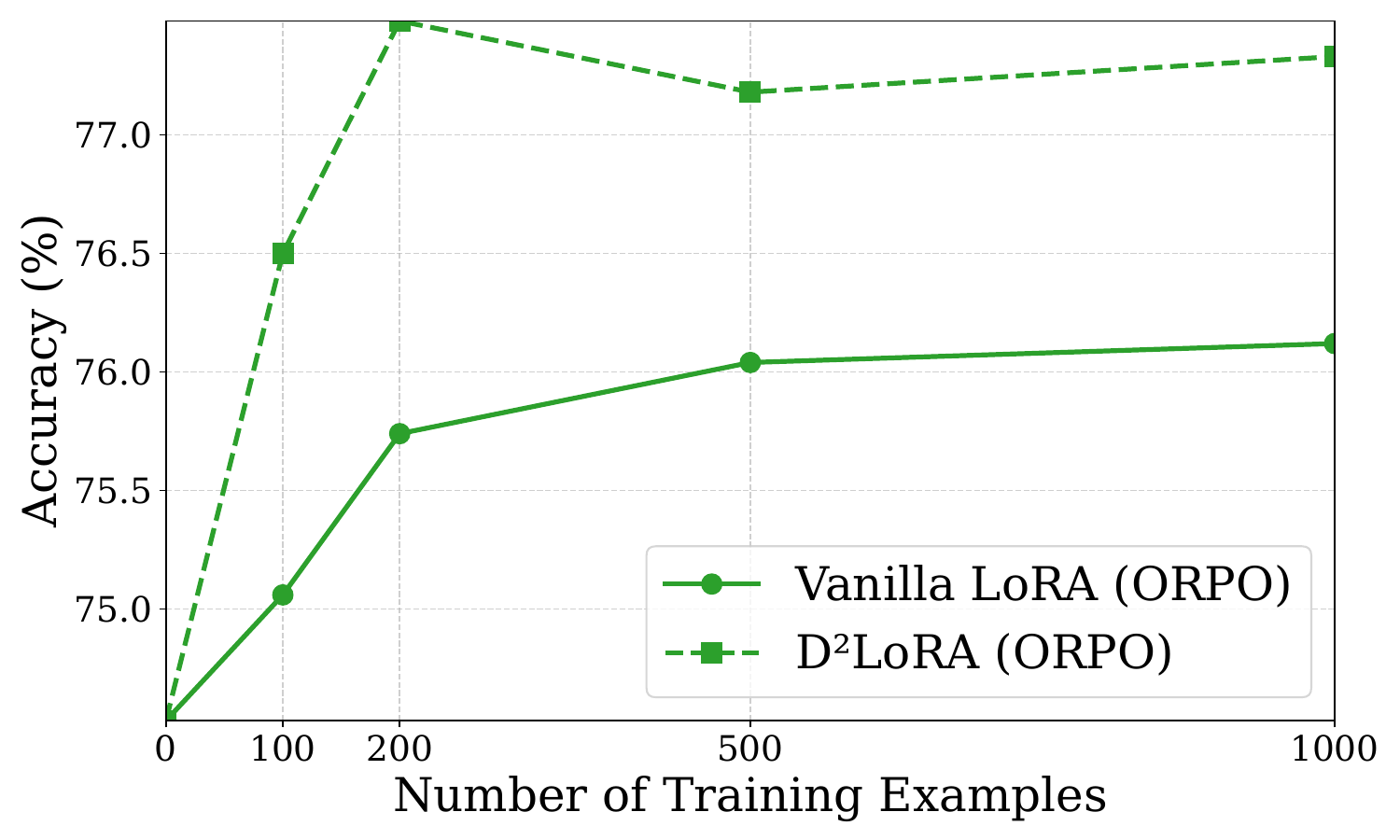}
    \caption{Comparison of accuracy between vanilla LoRA and \dtwolora on the GSM8K benchmark in a data-constrained setting, with training on 100, 200, 500, and 1000 samples.}
    \label{fig:less_1000}
\end{figure}

\begin{table}[ht]
    \centering
    \small 
    \resizebox{\columnwidth}{!}{
    \begin{tabular}{lccccc}
        \toprule
        \multirow{2}{*}{\textbf{Training Method}} & \multicolumn{5}{c}{\textbf{Number of Training Instances}} \\ 
        \cmidrule(lr){2-6}
        & \textbf{0} & \textbf{1k} & \textbf{2k} & \textbf{5k} & \textbf{10k} \\
        \midrule
        SFT               & 74.53 & 75.82 & 78.17 & 79.98 & 80.29 \\
        SFT (\dtwolora(40k,n))   & 74.96 & 77.10 & 79.53 & 80.44 & 80.67 \\
        \midrule
        ORPO             & 74.53 & 76.12 & 78.32 & 81.05 & 81.50 \\
        ORPO (\dtwolora(40k,n))    & 74.96 & 77.33 & 79.68 & 81.27 & 81.73 \\
        \midrule
        DPO               & 74.53 & 80.36 & 81.88 & 81.73 & 83.32 \\
        DPO (\dtwolora(40k,n))     & 74.96 & 80.82 & 82.94 & 82.87 & 83.47 \\
        \bottomrule
    \end{tabular}
    }
    \caption{Accuracy (\%) on the GSM8K benchmark for SFT, ORPO, and DPO, comparing vanilla LoRA with \dtwolora, which incorporates a warm-up phase with 40k samples}
    \label{tab:d2lora_math_results}
\end{table}

\end{document}